\documentclass[5p,times]{elsarticle}

\usepackage[english]{babel}

\usepackage{arabtex}
\def\endabstract{\egroup}

\usepackage{subfig}
\usepackage{graphicx}
\usepackage{multicol}
\usepackage{tikz}
\usepackage{amsmath}
\usepackage{bm}
\usepackage[export]{adjustbox}
\usepackage{color, soul}
\usepackage[
  separate-uncertainty = true,
  multi-part-units = repeat
]{siunitx}

\usepackage{float}
\usepackage{utf8}
\usepackage{comment}
\usepackage{verbatim}
\usepackage{booktabs}
\usepackage{appendix}

\DeclareFontFamily{U}{xnsh}{}%

\DeclareFontShape{U}{xnsh}{m}{n}{%
   <-6> sfixed * [6.0] xnsh14
      <6-10> s * [1.20] xnsh14
         <10><10.95><12><14.4><17.28><20.74><24.88> s * [1.0] xnsh14
         }{}

\DeclareFontShape{U}{xnsh}{bx}{n}{%
   <-6> sfixed * [6.0] xnsh14bf
   <6-10> s * [1.20] xnsh14bf
   <10><10.95><12><14.4><17.28><20.74><24.88> s * [1.20] xnsh14bf
}{}

\makeatletter
\def\ps@pprintTitle{%
  \let\@oddhead\@empty
  \let\@evenhead\@empty
  \let\@oddfoot\@empty
  \let\@evenfoot\@oddfoot
}
\makeatother

\usetikzlibrary{shapes.geometric, arrows}
\tikzstyle{startstop} = [rectangle, rounded corners, minimum width=3cm, minimum height=1cm,text centered, draw=black, fill=red!30]
\tikzstyle{io} = [trapezium, trapezium left angle=70, trapezium right angle=110, minimum width=3cm, minimum height=1cm, text centered, draw=black, fill=blue!30]
\tikzstyle{process} = [rectangle, text width= 6cm, minimum height=1cm, text centered, draw=black,fill=white!30]
\tikzstyle{decision} = [diamond, minimum width=3cm, minimum height=1cm, text centered, draw=black, fill=green!30]
\tikzstyle{arrow} = [thick,->,>=stealth]
\usepackage{multirow}
\usetikzlibrary{backgrounds,calc,positioning}

\usepackage{xcolor}

\usepackage{rotating}
\usepackage{array}
\usepackage{epstopdf}
\AppendGraphicsExtensions{.tif}

\usepackage{hyperref}
\hypersetup{
    colorlinks=true,
    linkcolor=blue,
    filecolor=magenta,      
    urlcolor=cyan,
    pdftitle={Overleaf Example},
    pdfpagemode=FullScreen,
    }

\begin{document}
\begin{frontmatter}
\title{Buildings Classification using Very High Resolution Satellite Imagery}
%\title{Road Accidents Scaling Analysis based on Social Media Crowdsourced Data\vspace{2cm}}

%\begin{comment}

\author[First]{Mohammad Dimassi}
\author[First]{Abed Ellatif Samhat}
\author[Second]{Mohammad Zaraket}
\author[Second]{Jamal Haidar}
\author[Third]{Mustafa Shukor}
\author[Fourth]{Ali J. Ghandour\corref{cor1}}
\address[First]{CRSI, Faculty of Engineering, Lebanese University | \{md22.dimassi@gmail.com, samhat@ul.edu.lb\}}
\address[Second]{CCE, Faculty of Engineering, Islamic University of Lebanon | \{mohzrkt99@gmail.com, jamal.haydar@iul.edu.lb\}}
\address[Third]{MINES ParisTech, Paris, France | mustafa.shukor@mines-paristech.fr}
\cortext[cor1]{Corresponding Author.\\Email: aghandour@cnrs.edu.lb}
\address[Fourth]{National Center for Remote Sensing - CNRS, Beirut, Lebanon | aghandour@cnrs.edu.lb}

%\end{comment}

\begin{abstract}
Buildings classification using satellite images is becoming more important for several applications such as damage assessment, resource allocation, and population estimation. We focus, in this work, on buildings damage assessment (BDA) and buildings type classification (BTC) of residential and non-residential buildings. We propose to rely solely on RGB satellite images and follow a 2-stage deep learning-based approach, where first, buildings' footprints are extracted using a semantic segmentation model, followed by classification of the cropped images. Due to the lack of an appropriate dataset for the residential/non-residential building classification, we introduce a new dataset of high-resolution satellite images. We conduct extensive experiments to select the best hyper-parameters, model architecture, and training paradigm, and we propose a new transfer learning-based approach that outperforms classical methods. Finally, we validate the proposed approach on two applications showing excellent accuracy and F1-score metrics.
\end{abstract}

\begin{keyword}
Deep learning \sep CNN \sep Transfer Learning \sep Buildings Type Classification \sep Building Damage Classification.
\end{keyword}
\end{frontmatter}

\section{Introduction}

\label{Introduction}
Buildings classification is vital for many applications, such as buildings damage assessment (BDA) and buildings type classification (BTC). Urban areas are constantly struck by man-made and/or natural disasters, such as wars, tornadoes, and earthquakes, resulting in large-scale buildings and urban infrastructure destruction. In the early reconstruction phase, damage assessment is conducted manually using crucial information about the area, amount, rate, and type of damage. In addition, buildings type classification (\emph{e.g.}, residential/non-residential) pave the way for many real-world applications such as population estimation and resource allocation.

For these purposes, remote sensing techniques can play an important role, mainly due to their wide availability at relatively low cost, wide field of view, and fast response capacities. Using deep learning for building classification can speed up the process by reducing human intervention and saving considerable time and cost. Indeed, we witnessed in past years a rapid development in the field of deep learning and its applications in earth observation, remote sensing, and computer vision fields \cite{DeepLearningInRS}.  

Contrary to other work that proposes to tackle such task by adopting one model for semantic segmentation with many classes \cite{iglovikov2017satellite, guerin2021satellite}, here we follow a 2-stage approach where we disentangle the semantic segmentation from the classification. In a nutshell, a semantic segmentation model takes an input RGB image and predicts buildings' masks (\emph{i.e.}, stage 1). In the second stage, buildings are cropped from the original image and fed to a classification model to predict the class of each building. In addition, we propose to use only RGB satellite images, which is more efficient than using additional other modalities.

This paper focuses on the second classification stage and conducts extensive experiments to find the best hyper-parameter, training paradigms, and model architecture for the underlying task. Moreover, we propose a new transfer learning paradigm that extends the classical 2-stage approach (\emph{i.e.}, pre-training then fine-tuning) by an additional stage that makes the model's layers more consistent and specific to the task, and we show that this outperforms the classical approach. We validate the proposed approach on two main tasks (\emph{i.e.}, BDA and BTC), showing excellent performance in terms of accuracy and F1-score metrics. Finally, we propose a new dataset for BTC. To avoid redundancy, we optimized the hyperparameters and the training paradigm for BDA while we focused on the architecture for BTC.
The list of contributions are the following:
\begin{itemize}
    \item We propose to use only RGB satellite images for buildings classification. We validate the approach on two main tasks; Buildings type classification (BTC) and Buildings damage assessment (BDA).
    \item We propose a new dataset for BTC.
    \item We conduct extensive experiments to choose the best hyperparameters, training paradigm, and model architecture.% for this specific task.
    \item We propose a new transfer learning paradigm that outperforms the classical 2-stage approach.
\end{itemize}

The rest of this paper is organized as follows: 
Section \ref{sec:related_work} reviews some of the related works. Section \ref{sec:method} presents a brief background and details the methodology adopted in this work. In Sections \ref{sec:app_bda} and \ref{sec:app_btc}, we validate the proposed approach using Building Damage Assessment and Building Damage Assessment applications, respectively, and discuss the proposed dataset. Finally, Section \ref{Conclusion} concludes this manuscript.

\section{Related Work}
\label{sec:related_work}

\subsection{Buildings Damage Assessment}
A vision-based approach for detecting cracks on concrete images is discussed in \cite{3} using deep learning techniques. Based on Convolution Neural Network (CNN), the idea is to determine cracks in a specific zone such as a building, especially in the roof, and make decisions that help classify the building as either damaged or not.
Authors in \cite{9} assess the impact of the combined use of different resolution satellite images on improving classification accuracy of damaged buildings using CNN.
In \cite{4},  the proposed method also detects Flooded/Damaged buildings from satellite imagery of an area affected by a hurricane. 
The authors prepared their dataset from online sources and applied several pre-processing steps before the training phase on the TensorFlow framework.
The study in \cite{10} proposes an algorithm for building damage detection from post-event aerial imagery, using a data expansion Single Shot multi-box Detector (SSD) algorithm for a small data set of Hurricane Sandy. The proposed algorithm relies on Feed Forward Neural Network (FFNN) and uses VGG-16 \cite{Vgg16} as the primary network to extract feature information.
In \cite{11}, the authors use CNN on a small set of candidates damaged buildings to reduce needed processing time.
Deep learning is also used in \cite{17} to improve the detection of rooftop hail damage. 
One CNN classifier is trained from scratch, while the second classifier relies on the set features from a pre-trained network.
In addition, different input size images were tested to check the best one. 

Moreover, researchers from various disciplines such as civil engineering, infrastructure, and mechanical engineering developed CNN models to detect different types of damage. The authors in \mbox{\cite{MEC1} and \cite{MEC2}} generate damage features' map using data extracted from sensors.  In \mbox{\cite{MEC1}}, authors provide structural damage localization with good accuracy using both noise-free and noisy datasets. In \mbox{\cite{MEC2}}, as the variation of the temperature alters the structural model parameters, a  damage detection technique is proposed to consider both uncertainties and varying temperatures. It is developed on the basis of Sparse Bayesian Learning (SBL). 
An application for an unsupervised learning approach was presented in \mbox{\cite{UNSUP}} where the authors propose a transfer learning approach to use a source network trained on a labeled dataset to be able to train the target network on an unlabeled dataset.

\begin{figure*}[t]
    \centering
    \includegraphics[width=\textwidth]{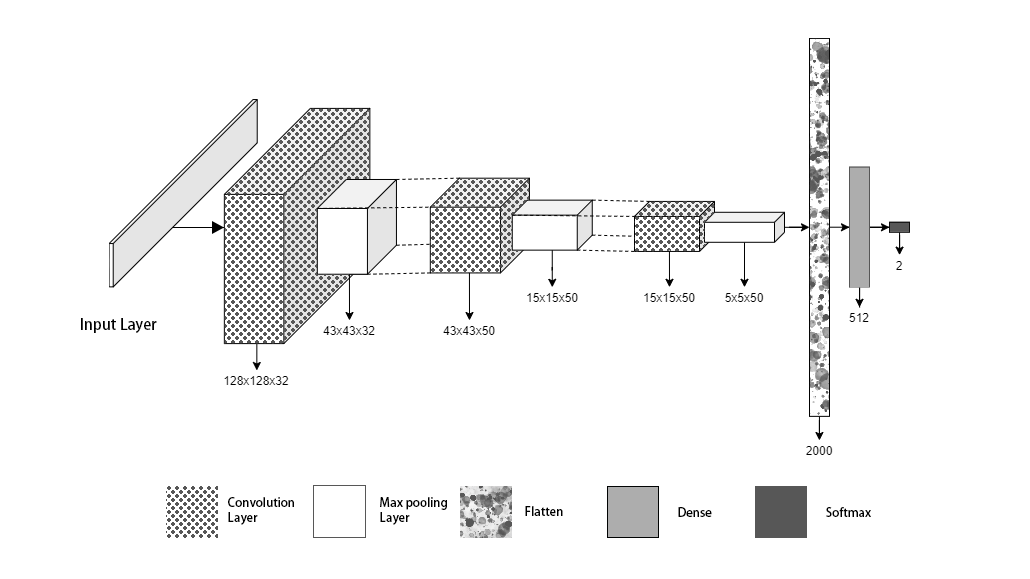}
    \caption{Architecture of the adopted CNN network model for BDA application.}
    \label{fig: CNN}
\end{figure*}

In all previous work in the literature related to damaged buildings, the optimization of the hyper-parameters for the deep learning model was barely investigated.  

\subsection{Buildings Type Classification}
In \cite{sturrock2018predicting}, authors use an ensemble of machine learning models to classify buildings as sprayable and not-sprayable based on buildings characteristics such as size, shape, and proximity to neighboring features. Similarly, in \cite{GISML}, authors use classical machine learning approaches to predict if the building is residential or not based on several input variables (stored in shapefiles, CSV ...).

Work presented in \cite{nlp_china} propose to classify buildings type using geospatial data (\emph{e.g.},  point-of-interest (POI) data, building footprints, land use polygons, and roads) based on NLP and ratio-based techniques. In \cite{SpatiotempClus}, an iterative clustering method to classify buildings based on spatiotemporal data (\emph{e.g.}, population density and people interaction) is introduced.
Random Forest Classifier is used in \cite{AutoTopo} to classify buildings' footprint from different data sources (e.g., topographic raster maps, cadastral databases, or digital landscape models). Authors in \cite{ontoML} use object-Based Image Analysis (OBIA) and machine learning methods to extract and classify buildings from Airborne Laser Scanner (ALS). Gaussian finite mixture model is proposed in \cite{MSSPat} to classify buildings based on several metrics extracted from high-resolution satellite images. Finally, \cite{kang2018building} combines street view with satellite images to classify buildings using CNN models. 

To the best of our knowledge, no method in the literature is solely based on RGB aerial images to classify buildings types.

\section{Methodology}
\label{sec:method}
The primary motivation behind adopting the 2-stages approach is to fragment the main problem into two more straightforward tasks: semantic segmentation and image classification. This approach helps to leverage the recent advances in those two domains to solve the underlying task. In addition, besides being more efficient and cheaper, we argue that RGB images are enough to classify buildings, as this can be done relatively easily by humans.\\

In this section, we will focus on the second stage, which is buildings classification. In a nutshell, the model takes an RGB image of a building cropped using the predicted segmentation mask from the first stage and output the class of the building, either damaged or not and either residential or not.

\begin{figure*}[t]
    \centering
    \subfloat[]
        {
        \includegraphics[width=0.2\textwidth]{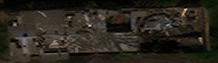}
        }    
    %\subfloat[]
    %    {
    %    \includegraphics[width=0.2\textwidth]{Figures/Damage5.jpg}
    %    }    
    \subfloat[]
        {
        \includegraphics[width=0.2\textwidth]{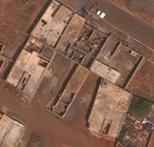}
        }    
    \subfloat[]
        {
        \includegraphics[width=0.2\textwidth]{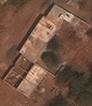}
        }
    \caption{Sample damaged buildings scenes from the xView dataset.}
    \label{fig: Damage Images}
\end{figure*}

\subsection{Loss Function}
Multiple loss functions exist and can be used during training. In this work, we focus on two of the widely used losses in order to compare their performance and choose the appropriate one for the task: (\textit{i}) Cross-Entropy loss and (\textit{ii}) Focal loss \cite{FocalLoss}.

Cross-Entropy loss is defined in Equation \ref{eq:CrossEntropy}, where $t_{i}$ and $s_{i}$ are, respectively, the ground-truth and the score for each class $i$ in the universe of classes $C$.
\begin{equation}
CE = -\sum_{i}^{C}t_{i} log (s_{i})\
\label{eq:CrossEntropy}
\end{equation}
In our case, we are dealing with a binary classification which means that $C = 2$, and this is referred to as the Binary Cross Entropy defined in Equation \ref{eq:BinaryCrossEntropy}.
$t_{1}$ and $s_{1}$ are, respectively, the ground-truth and the score for class $C_{1}$.

\begin{equation}
BCE = -\sum_{i=1}^{C=2}t_{i} log (s_{i}) = -t_{1} log(s_{1}) - (1 - t_{1}) log(1 - s_{1})
\label{eq:BinaryCrossEntropy}
\end{equation}

Focal loss \cite{FocalLoss} weighs the contribution of each sample to the loss based on the classification error. If a sample is classified correctly by the CNN, its contribution to the loss model should decrease. With this strategy, focal loss solves the problem of hard labels and plays a vital role in imbalanced classes dataset.
Binary focal loss is described in Equation \ref{eq:Focal}, where $(1-s_{i})$ $\gamma$ is a modulating factor to reduce the influence of correctly classified samples in the loss and $\gamma \geq 0$ is referred to as the focusing parameter. With $\gamma = 0$, focal loss is reduced to binary cross entropy. 
\begin{equation}
FL = -\sum_{i=1}^{C=2}(1 - s_{i})^{\gamma }t_{i} log (s_{i})
\label{eq:Focal}
\end{equation}

\subsection{Optimizer}
During CNN training, the role of the optimizer is to update the weight parameters to minimize the loss function. Multiple optimizers are suggested in the literature, such as Momentum, Regular Gradient Descent (RGD), Stochastic Gradient Descent (SGD), Adam, and RectifierAdam for classification tasks \cite{Optimizers}.

Although non-adaptive optimizers, such as SGD,  help obtain better minima and generalization properties. Adam optimizer is widely used as it leads to faster convergence due to its adaptive learning rate. However, Adam suffers from significant variance at the beginning of training.
Rectifier-Adam (also known as RectAdam) \cite{rectadam} was proposed to improve the convergence of Adam. 

In a nutshell, no optimizer works best for all the applications; thus, we propose a comparison of these optimizers in this work. 

\begin{figure*}[t]
    \centering
    \subfloat[]
        {
        \includegraphics[width=0.4\textwidth]{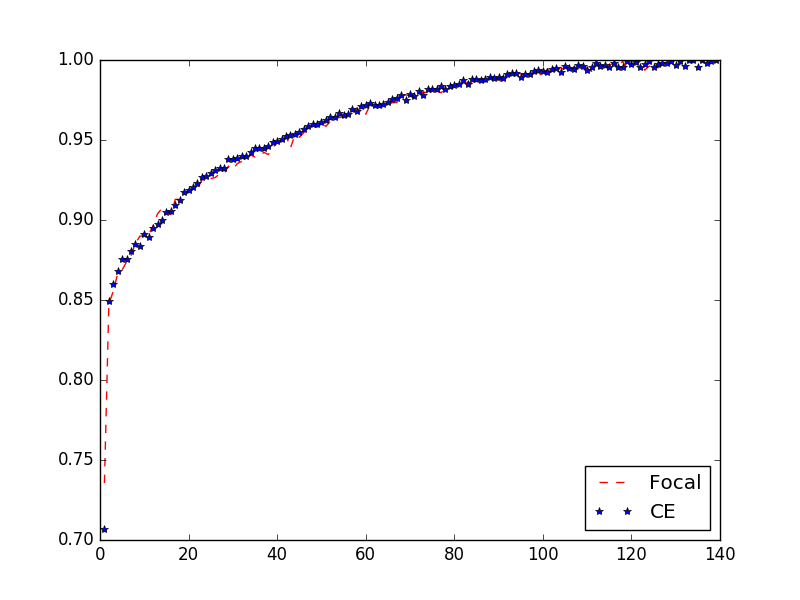}
        \label{fig: Acc of Losses Funcions}
        }    
    \subfloat[]
        {
        \includegraphics[width=0.4\textwidth]{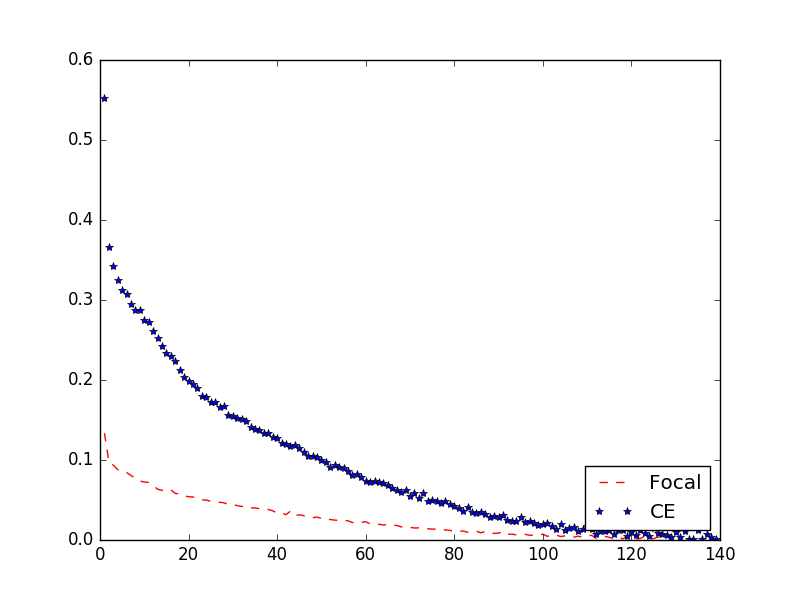}
        \label{fig: Loss of Losses Functions}
        }
    \caption{(a) and (b) show the accuracy and loss, respectively, in function of time for BDA: the model is trained using Cross Entropy and Focal Loss functions. From the loss curves, we can notice that the model using Focal Loss converges faster.}
    \label{fig: Losses Functions}
\end{figure*}

\subsection{Transfer Learning}
\label{TLsec}
Transfer learning is considered an innovative approach in CNN training to achieve high accuracy with minimal time and effort, relying on existing pre-trained model. The idea is to use a model trained on a large dataset and transfer its knowledge to the application at hand. 
The classical transfer learning approach adopts 2-stage strategies: (\textit{i}) a model is trained from scratch on a large dataset (\emph{i.e.}, pre-training on ImageNet \cite{imagenet}), then (\textit{ii}) the last layers of the model, are retrained for the downstream application.
Here, we propose to extend this approach by adding a third and final stage: (\textit{iii}) we freeze the last layers of the network and retrain initial layers (previously froze) in order to ensure the consistency between those layers and the last ones, and also to adjust all the weights according to the specific target application.

\subsection{Evaluation and Metrics}
\label{sec:metrics}
 We are relaying on the classical metrics used in this context which are: 
\begin{itemize}
\item  True positive (TP): Both manual and automated methods label the object belonging to the
buildings regions.
\item  True negative (TN): Both manual and automated methods label the object belonging to the background.
\item  False positive (FP): The automated method incorrectly labels the object as belonging to the building regions.
\item  False negative (FN): The automated method does not correctly label a pixel truly belonging to the
building regions.
\end{itemize}
These global definitions can be used to generate the following performance metrics defined in Equations (\ref{eq:Recall}), (\ref{eq:Precision}) and (\ref{eq:F1-score}).

\begin{equation}					
    Recall 
  = \dfrac{TP}{TP + FN}
    \label{eq:Recall}
\end{equation}

\begin{equation}
    Precision
  = \dfrac{TP}{TP + FP}
    \label{eq:Precision}
\end{equation}

\begin{comment}
\begin{equation}					
    F-Score 
    = \dfrac{2\times TP}{2\times TP + FN + FP}
    \label{eq:F-score}
\end{equation}
\end{comment}

\begin{equation}					
    F1-Score 
    = \dfrac{2\times Pr \times Rc}{Pr + Rc}
    \label{eq:F1-score}
\end{equation}
Also one of the essential metrics is the accuracy given 
%, which is representative as follow 
in equation (\ref{eq:Accuracy})

\begin{equation}					
    Accuracy 
    = \dfrac{TP + TN}{TP + TN + FP + FN}
    \label{eq:Accuracy}
\end{equation}

In addition, we report the speed of convergence for each scenario using the hyper-parameter "number of epochs" taken by the network to converge.

\section{Buildings Damage Assessment}
\label{sec:app_bda}

The following two sections detail how the proposed approach can be applied to BDA and BTC applications. For BDA, we focus on the choice of the training paradigm (\emph{e.g.}, Transfer Learning). For BTC, we focus on the architecture.\\

\begin{figure*}[t]
    \centering
    \subfloat[]
        {
        \includegraphics[width=0.4\textwidth]{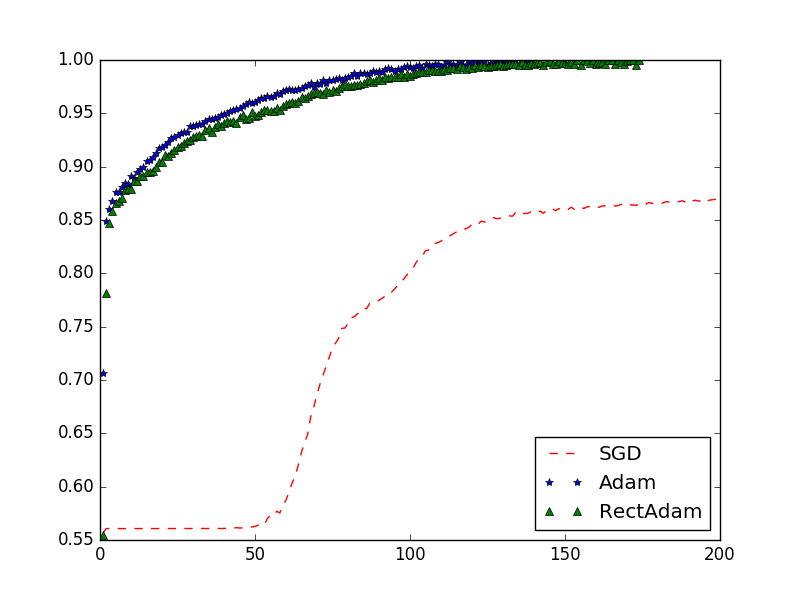}
        \label{fig: Acc of Optimizers}
        }    
    \subfloat[]
        {
        \includegraphics[width=0.4\textwidth]{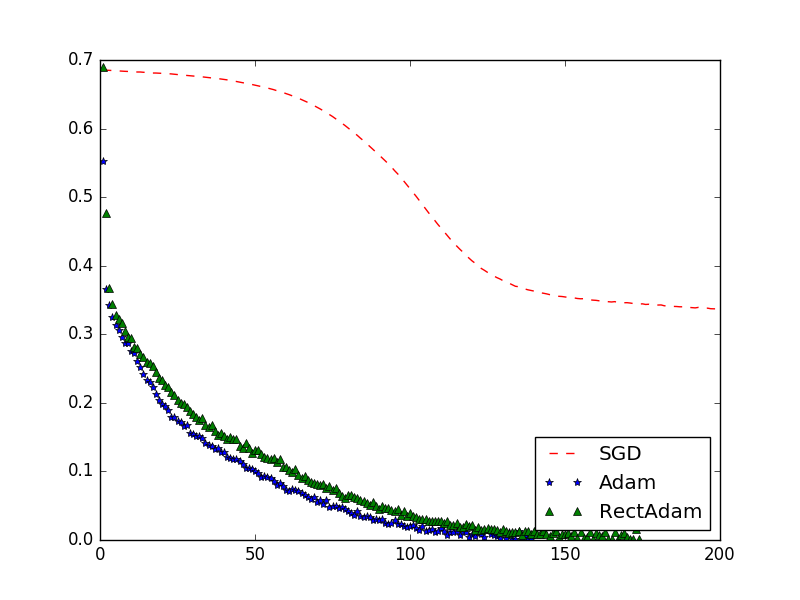}
        \label{fig: Loss of Optimizers}
        }
    \caption{(a) and (b) show the accuracy and loss, respectively, in function of time for BDA: the model is trained with SGD, Adam and RectAdam optimizers. The model trained with SGD struggle to converge. Adam optimizer leads to faster convergence.}
    \label{fig: Optimizers}
\end{figure*}

\begin{table*}[t]
\setlength\tabcolsep{2pt}%
\centering
\small
\begin{tabular}{ccccc}
\toprule
Dataset & \begin{tabular}{c} Number \\ of Objects   \end{tabular} & \begin{tabular}{c} Resolution \\ (in pixels)   \end{tabular} & Format & Generalizable\\ \midrule
ABCD dataset  \cite{45} &  8,500   & 128x128  & .tif &  No\\ 
Flooding dataset \cite{4} &  10,000 & 128x128  & .jpeg & No \\ 
NOAA dataset \cite{NOAA} & 500,000 & 9351x9351 & .tif & No \\ 
xView dataset \cite{xView} & 1,000,000 & 3000x3000 & .tif & Yes\\ 
\bottomrule
\end{tabular}
\caption{Datasets survey and comparison.}
\label{dataset}

\end{table*}

Buildings Damage Assessment task consists of classifying buildings as damaged or not, in which a classification model takes an RGB image and outputs the image class. This section gives an overview of existing datasets and then explains the implementation details and experimental results.

\subsection{Dataset}
\label{Dataset Survey}

Dataset choice is an essential component of our work. CNN relies on labeled data to train its network and provides a high-accuracy classifier model. Four datasets are considered, analyzed, and compared to choose the appropriate one for our research.

We used several parameters for the datasets benchmark. Damage type was the critical metric for choosing the appropriate dataset. Damage type differs based on the disaster event. War, earthquakes, and wind (hurricanes and tornadoes) usually result in wholly or partly demolished and ruined buildings. Flooding produces water in areas where it is not normally expected, while volcano erupts lava.    

For the scope of this work, we define a damaged building as a partly or wholly demolished building. This definition covers the most common damage type resulting from armed conflicts, earthquakes, tornadoes, and hurricanes. 

We considered four different datasets as potential candidates for the BDA application: ABCD Dataset \cite{45}, Flooding Dataset \cite{4}, NOAA Dataset \cite{NOAA} and xView Dataset \cite{xView}. Table \ref{dataset} shows a comparison between these four datasets. One can conclude that xView dataset is the best candidate for our research. xView dataset is well labeled and is big enough so that we will choose it for BDA task. Figure \mbox{\ref{fig: Damage Images}} shows sample damaged scenes from the xView dataset.

\paragraph{xView Dataset}
 xView is one of the largest publicly available datasets with annotated images from complex scenes around the world.
xView dataset contains more than 1 million labeled objects covering over 1400 $Km^2$. Images have an average of (3000x3000) pixels resolution using the “.tif” format and high-quality labeling, which provides a large amount of available imagery to understand the visual world in new ways and address a range of applications, including damaged building classification.

\subsection{Implementation}
For each scenario in the following subsections, we rely on a CNN network made from three layers of Convolutions + MaxPooling followed by Flatten, Dense, and finally, Softmax layer. The used network is depicted in Figure \mbox{\ref{fig: CNN}}. 
In our work, we used the Sigmoid activation function for each convolution layer and the ReLU activation function for the dense layer. 

We carried an evaluation study over the test data. The convergence condition is to reach a high accuracy value (~99\%) and low loss threshold ($\leq$ 0.001) over the trained data.

\subsection{Experimental Results}
\label{Results}

In this section, we present the results for the different scenarios discussed in Section \ref{sec:method} based on the xView dataset.

\paragraph{Loss Function Results}
In order to choose the best loss function suitable for the damaged building application, we built two models using the architecture described in Figure \mbox{\ref{fig: CNN}}, where the first one uses cross-entropy loss function and the second one uses focal loss function. 
After training the two models corresponding to cross-entropy loss and focal loss until converging, we present the results during the training phase and for the test dataset in Figure \ref{fig: Losses Functions} and Table \ref{table: losses}.
Figures \ref{fig: Losses Functions} (a) and (b) show the accuracy and the loss as a function of time (number of epochs) for Cross-Entropy and Focal Loss functions.
Figure \ref{fig: Acc of Losses Funcions} shows clearly that both loss functions converge to 99\% accuracy over the trained data, but focal loss converges faster during training as shown in Figure \ref{fig: Loss of Losses Functions}. In addition, Table \ref{table: losses} confirms that focal loss converges faster than cross-entropy (127 epochs versus 138 epochs). Finally, we can notice that focal loss scores slightly higher accuracy than cross-entropy over the test dataset. \\
To conclude, focal loss has better results against cross-entropy loss in terms of the following three metrics: accuracy, number of epoch to converge, and loss per epoch. Thus, we deduce that focal loss has a better performance for damaged building classification scope, and thus it will be adopted as part of the following scenario.

\begin{table}[t]
\setlength\tabcolsep{2pt}%
\centering
\small
\begin{tabular}{ccc}
\toprule
 & Epochs & Accuracy (\%) \\ \midrule
Cross Entropy Loss & 138 & 97.33 \\ 
Focal  Loss & \bf{127} & \bf{97.56} \\ 
\bottomrule
\end{tabular}
\caption{The model is trained on Cross Entropy and Focal Loss functions for BDA. We can notice that the model with Focal Loss converges faster.}
\label{table: losses}

\end{table}

\paragraph{Optimizer Results}
No single optimizer choice can be made for all different CNN applications. For instance, Adam optimizer is widely adopted and used for several applications, but still, in many use cases, SGD has reported better results. RectAdam \cite{rectadam} was introduced as an enhancement for Adam. However, in multiple scenarios, Adam has shown better and faster performance.

Accordingly, and for the scope of this work,  we  consider three optimizers:  (\textit{i}) SGD, (\textit{ii}) Adam and (\textit{iii}) RectAdam. We will compare their suitability for damaged building classification application and select the best option.

The results for SGD, Adam, and RectAdam optimizers during the training phase and for the test dataset using xView, are presented in Table \ref{table: Optimizers} and Figure \ref{fig: Optimizers}.
Figures \ref{fig: Optimizers}a and b show the accuracy metric and the loss metric, respectively, as a  function of time (number of epochs) for SGD, Adam, and RectAdam optimizers.

\begin{table}[t]
\setlength\tabcolsep{2pt}%
\centering
\small
\begin{tabular}{ccc}
\toprule
 & Epochs & Accuracy (\%) \\ \midrule
SGD & - & 85.05 \\ 
Adam & \bf{138} & \bf{97.33} \\ 
RectAdam & 174 & 97.23 \\ 
\bottomrule
\end{tabular}
\caption{The model is trained with different optimizers for BDA; we can notice that Adam leads to faster convergence while SGD struggle to converge.}
\label{table: Optimizers}

\end{table}

Results in Figure \ref{fig: Optimizers} reveal that SGD struggled to converge and was not able to reach the desired accuracy and loss thresholds during training, whereas Adam and RectAdam optimizers both converge rapidly.

In terms of accuracy, Adam and RectAdam score better result (97.3\% and 97.2\%, respectively) than SGD (85.1\%) as tabulated in Table \ref{table: Optimizers}. Regarding the number of epochs needed to converge, Adam is much faster than RectAdam (137 epochs vs. 174 epochs).

These results clearly show that Adam and RectAdam have better performance than SGD in terms of accuracy, the number of epochs to converge, and loss per epoch. We will adopt Adam optimizer for damaged building classification application in the following scenarios.

\paragraph{Transfer Learning Results}
Our goal is to apply multiple models based on Transfer Learning (TL) and find the best methodology for training our damaged building classifier using TL. For this sake, we consider the following three models: 
\begin{itemize}
\item Model 1: network trained from scratch without using any pre-trained weights. We will refer to this model as \textbf{Baseline} model.
\item Model 2: network based on transfer learning approach using VGG16 pre-trained weights on ImageNet dataset \cite{Vgg16} where we freeze all layers and train only the last layers of the network. We will refer to this model as transfer learning - VGG16 (\textbf{TL-VGG16}) model.
\item Model 3: enhanced transfer leaning approach (discussed in Section \ref{TLsec}) where an additional final step is added. We will refer to this model as enhanced - transfer learning - VGG16 (\textbf{E-TL-VGG16}) model.
\end{itemize}

In this scenario, the three models above are compared and analyzed.
Table \ref{Table: TL} clearly shows the difference in speed convergence, where TL-VGG16 iterates only for 23 epochs to converge, while the Baseline model needed 127 epochs.

From Table \ref{Table: TL}, one can notice that TL-VGG16 model converges to the desired thresholds faster than the Baseline model. Since using pre-trained weights would help the network to converge faster. However, this does not necessarily result in higher accuracy, as shown in Table \ref{Table: TL}. TL-VGG16 achieved 96\% accuracy over the test data, which is lower than the 97.5\% scored by the Baseline model.

The rationale behind E-TL-VGG16 model is to improve the accuracy of the transfer learning approach while maintaining good enough convergence speed.
At epoch 23, when we un-freeze the VGG16 layers and de-freeze the last layers, accuracy metric sharply decreases, and loss sharply increases. After this glitch in performance, the network spends 69 epochs to re-converge to previously attained thresholds.

E-TL-VGG16 performance tabulated in Table \ref{Table: TL} reveals an important improvement in the last percentages of the accuracy to reach 98.96\%. Finally, we also compute the F1-score achieved by E-TL-VGG16, which turns to be equal to 99.4\%.

\begin{table}[t]
\setlength\tabcolsep{2pt}%
\centering
\small
\begin{tabular}{ccc}
\toprule
 & Epochs & Accuracy (\%) \\ \midrule
Baseline & 127 & 97.33 \\ 
TL-VGG16 & \bf{23} & 96.12 \\ 
E-TL-VGG16 (ours) & 69 (23 + 46) & \bf{98.96} \\ 
\bottomrule
\end{tabular}
\caption{Comparison of different trained paradigms for BDA; ours (E-TL-VGG16) is the best in terms of accuracy and faster to train than the baseline.}
\label{Table: TL}

\end{table}

\section{Buildings Type Classification}
\label{sec:app_btc}
This section elaborates on the approach used to classify buildings into residential or non-residential using only RGB aerial images. This task can pave the way for several applications, such as population estimation and resource allocation. Due to the lack of an appropriate dataset for such a task, we introduce a new "Beirut Buildings Type Classification" (BBTC) dataset to help the community develop this field further. We start by introducing the dataset then explaining reported experimental results.

\subsection{BBTC Dataset}
Here, we propose a new dataset referred to as Beirut Buildings Type Classification (BBTC) after presenting an overview of existing datasets:
\paragraph{Related Datasets} To the best of our knowledge, there is no high-resolution satellite images dataset for classifying buildings into residential/non-residential. Among existing datasets, we found only DSTL dataset \cite{Dstl} that provides 1km x 1km satellite images in both 3-band and 16-band formats. However, DSTL images' resolution is not high (\emph{i.e.}, 1.24 m) and the dataset is highly imbalanced (\emph{i.e.}, 3.1\% non-residential buildings).

\paragraph{BBTC Description} The dataset consists of 17,033 RGB images, where each image contains a single building and is annotated as residential (15,330 buildings) or non-residential buildings (1,703 buildings which accounts for 10\% of annotated objects). The tiles are extracted from high resolution images (\emph{i.e.}, $\sim$ 50 cm) and covers Beirut city. Some examples can be seen in Figure \ref{fig:dataset_examples}.

\paragraph{BBTC Creation} The dataset was created and labeled using the following steps:
\begin{itemize}
    \item buildings polygons extracted from OpenStreetMap as a shapefile.

    \item annotation done using QGIS software; shapefile placed on two base maps for visual confirmation (Bing Virtual Earth and Google Satellite Hybrid) and each building labeled as residential or non-residential. Entire annotated buildings are almost 20,000 objects. Non-residential objects include universities, schools, governmental facilities, mosques, churches, commercial centers, and industrial facilities. 
    \item corresponding tile for each building requested from Mapbox at zoom level 17.
    \item finally, buildings cropped (slightly larger than the corresponding polygon to account for any inaccuracy) and saved. Duplicate buildings (\emph{i.e.}, retrieved multiple times) are ignored, reaching a total of 17,033 objects.

\end{itemize}

BBTC dataset can be accessed via the following bucket url:
\url{https://storage.googleapis.com/bbtc/bbtc\_dataset.tar.gz}

\begin{figure*}[t]
\setlength\tabcolsep{2pt}%
\centering
\begin{tabular}{ccccc|ccccc}
% \centering
\toprule
\includegraphics[width=0.09\textwidth]{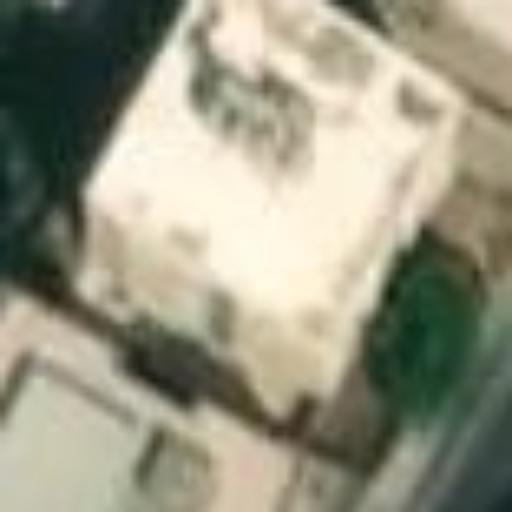}  & 
\includegraphics[width=0.09\textwidth]{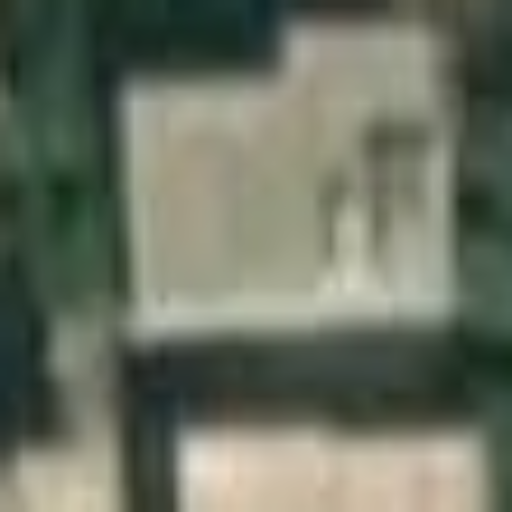}  & 
 \includegraphics[width=0.09\textwidth]{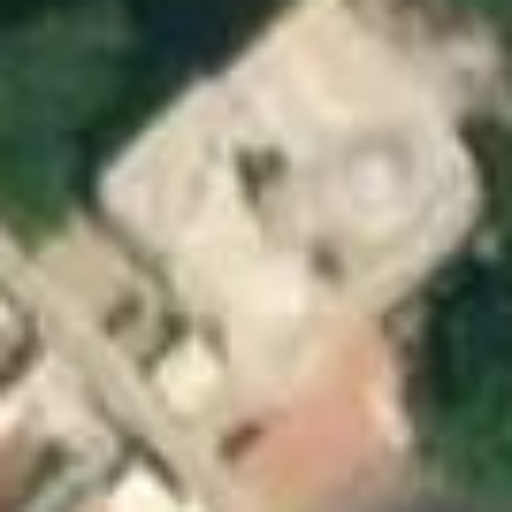}  & 
 \includegraphics[width=0.09\textwidth]{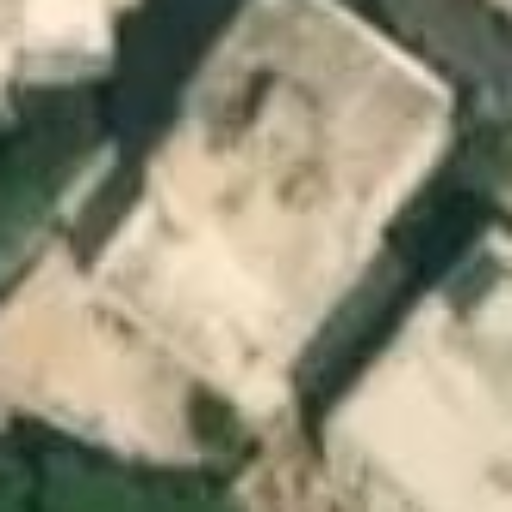}  & 
 \includegraphics[width=0.09\textwidth]{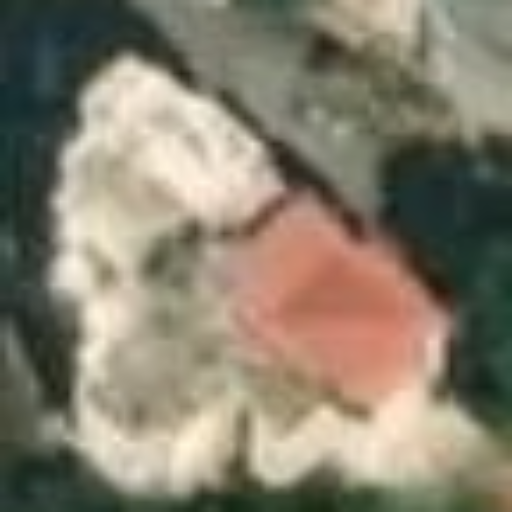}  & 
\includegraphics[width=0.09\textwidth]{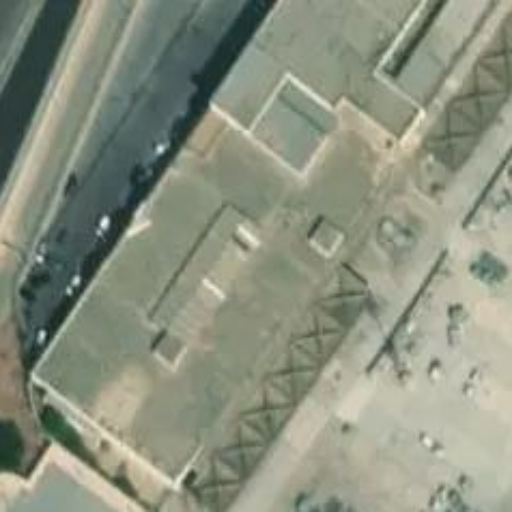}  & 
\includegraphics[width=0.09\textwidth]{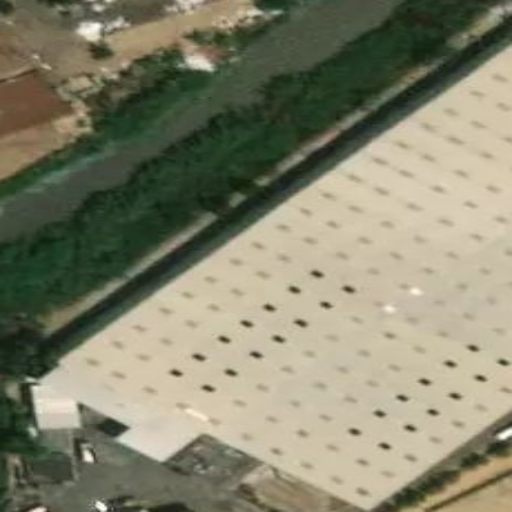}  & 
 \includegraphics[width=0.09\textwidth]{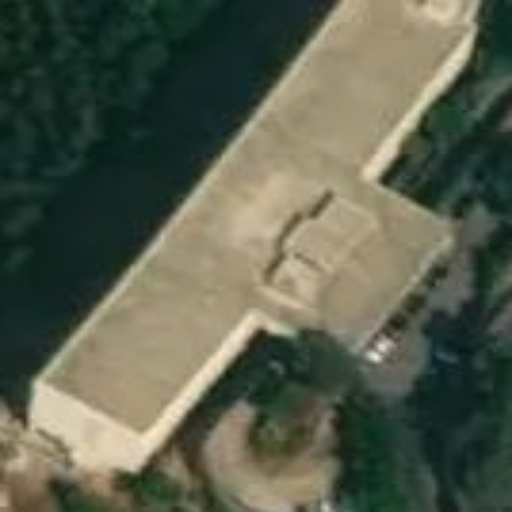}  & 
 \includegraphics[width=0.09\textwidth]{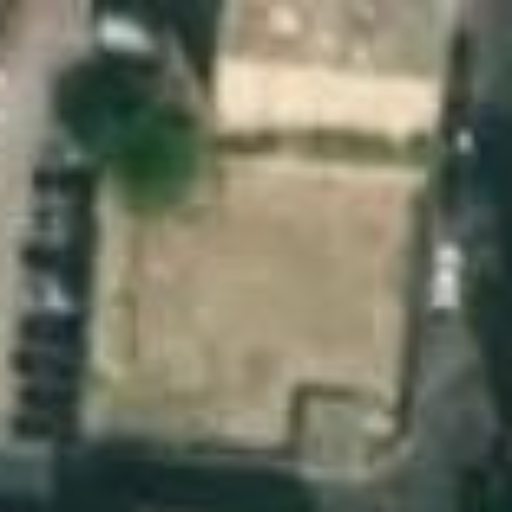}  & 
 \includegraphics[width=0.09\textwidth]{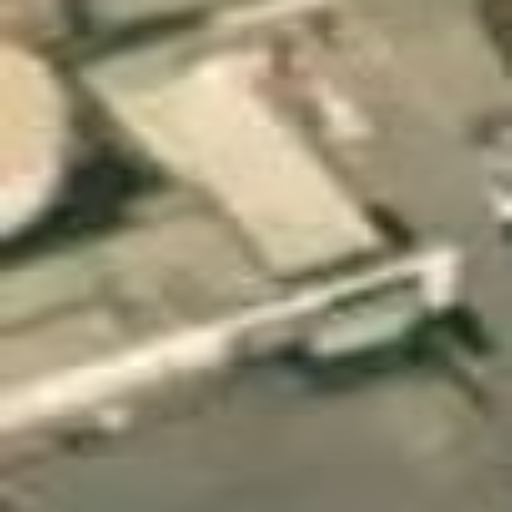}  \\
 \multicolumn{5}{c|}{Residential} & \multicolumn{5}{c}{Non Residential} \\
\bottomrule
\end{tabular}
\caption{Sample buildings' images from the proposed BBTC dataset.}
\label{fig:dataset_examples}
\end{figure*}

\subsection{Experimental Results}
This section provides and discusses experimental results for BTC. 

\paragraph{Architecture} 
We compared 5 architectures including state of the art ones; VGG16, ResNet50 \cite{resnet}, Res2Net \cite{res2net}, EfficientNet-b0 \cite{efficientnet} and RexNet \cite{rexnet}. All models trained with CE and SGD optimizers. Table \ref{tab:bta_archi} shows that RexNet gives the best results in terms of accuracy at the test set (94.8\%), while average time per epoch (3minutes 15 seconds) is almost equal to EfficientNet. This result does not come as a surprise since RexNet is a newly proposed architecture where the authors claim to provide high accuracy while achieving low time complexity comparable to EfficientNet.

\paragraph{Optimizer}
Finally, we adopt the winning model (\emph{i.e.}, RexNet) and train it using different optimizers. From Table \ref{tab:bta_optim}, one can notice that SGD optimizer was able to converge, in contrast to previous performance witnessed in Table \ref{table: Optimizers}. SGD achieved the best accuracy on the test set (94.8\%) compared to Adam and RectAdam, although it is the slowest with convergence time equal to 99 epochs.

\begin{table}[t]
\setlength\tabcolsep{2pt}%
\centering
\small
\begin{tabular}{ccccc}
\toprule
Model & \begin{tabular}{c}  Avg. time \\ (per epoch) \end{tabular} & Val. Acc & Epoch & Test Acc \\ \midrule
VGG16 & 3m20s & 0.953 & 95 & 0.935 \\
ResNet 50 & 5m30s & 0.947 & 44 & 0.933 \\
Res2Net & 9m30s & 0.960 & \bf{40} & 0.943 \\
EfficientNet & \bf{3m13s} & 0.960 & 49 & 0.941 \\
RexNet & 3m15s & \bf{0.962} & 99 & \bf{0.948} \\
\bottomrule
\end{tabular}
\caption{Ablation study for BTC on the proposed dataset; We compare different architectures. RexNet outperforms all the other models in terms of accuracy.}
\label{tab:bta_archi}
\end{table}

\begin{table}[t]
\setlength\tabcolsep{2pt}%
\centering
\small
\begin{tabular}{ccccc}
\toprule
Optimizer & \begin{tabular}{c}  Avg. time \\ (per epoch) \end{tabular} & Val. Acc & Epoch & Test Acc \\ \midrule

SGD & 3m15s & \bf{0.962} & 99 & \bf{0.948} \\
Adam & 3m15s & 0.957 & 90 & 0.945 \\
RectAdam & 3m15s & 0.958 & \bf{68} & 0.947 \\

\bottomrule
\end{tabular}
\caption{Ablation study for BTC on the proposed dataset; RexNet is trained with several optimizers. SGD leads to the best accuracy while RectAdam leads to faster convergence.}
\label{tab:bta_optim}
\end{table}

\section{Conclusion}
\label{Conclusion}

In this paper, we followed a 2-stage approach for buildings classification using only RGB satellite images. We focused here on the second stage, which consists of classification. We conducted extensive experiments to find the best hyperparameters, network architecture, and training paradigm, and we proposed a new transfer learning approach. Due to the lack of appropriate residential/non-residential buildings classification datasets, we proposed BBTC, a new dataset of high-resolution satellite images. We validated the proposed approach on two applications: Buildings Damage Assessment and Buildings Type Classification, which showed outstanding performance in several metrics on xView and BBTC datasets.

\section*{Conflict of Interest}
Declarations of interest: none

\bibliography{egbib}

\end{document}